\documentclass[twoside,11pt]{article}

\usepackage{jmlr2e}
\usepackage{amsmath}
\usepackage{times}
\usepackage{graphicx}
\usepackage{subfig}
\usepackage[shortlabels]{enumitem}

\DeclareMathOperator{\argmax}{arg\,max}
\DeclareMathOperator{\argmin}{arg\,min}

\ShortHeadings{Continuous State-Space Models for Optimal Sepsis Treatment}{Raghu, Komorowski, Celi, Szolovits, and Ghassemi}
\firstpageno{1}

\begin{document}

\title{Continuous State-Space Models for Optimal Sepsis Treatment - a Deep Reinforcement Learning Approach}

\author{Aniruddh Raghu \email araghu@mit.edu \\
       \AND
       \name Matthieu Komorowski \email mkomo@mit.edu\\
       \AND
       \name Leo Anthony Celi \email lceli@mit.edu \\
       \AND
       \name Peter Szolovits \email psz@mit.edu 		\\
       \AND
       \name Marzyeh Ghassemi \email mghassem@mit.edu \\
       \\
       \addr Computer Science and Artificial Intelligence Lab, MIT\\
       Cambridge, MA\\}

\maketitle

\begin{abstract}
Sepsis is a leading cause of mortality in intensive care units (ICUs) and costs hospitals billions annually. Treating a septic patient is highly challenging, because individual patients respond very differently to medical interventions and there is no universally agreed-upon treatment for sepsis. Understanding more about a patient's physiological state at a given time could hold the key to effective treatment policies. 
In this work, we propose a new approach to deduce optimal treatment policies for septic patients by using continuous state-space models and deep reinforcement learning. Learning treatment policies over continuous spaces is important, because we retain more of the patient's physiological information. 
Our model is able to learn clinically interpretable treatment policies, similar in important aspects to the treatment policies of physicians. Evaluating our algorithm on past ICU patient data, we find that our model could reduce patient mortality in the hospital by up to 3.6\% over observed clinical policies, from a baseline mortality of 13.7\%. The learned treatment policies could be used to aid intensive care clinicians in medical decision making and improve the likelihood of patient survival. 
\end{abstract}

\section{Introduction} 
Sepsis (severe infections with organ failure) is a dangerous condition that costs hospitals billions of pounds in the UK alone \citep{sepsiscost}, and is a leading cause of patient mortality \citep{sepsismortality}. The clinicians' task of deciding treatment type and dosage for individual patients is highly challenging. Besides antibiotics and infection source control, a cornerstone of the management of severe infections is administration of intravenous fluids to correct hypovolemia. This may be followed by the administration of vasopressors to counteract sepsis-induced vasodilation. Various fluids and vasopressor treatment strategies have been shown to lead to extreme variations in patient mortality, which demonstrates how critical these decisions are \citep{waechter2014interaction}. While international efforts attempt to provide general guidance for treating sepsis, physicians at the bedside still lack efficient tools to provide individualized real-term decision support \citep{rhodes2017surviving}. As a consequence, individual clinicians vary treatment in many ways, e.g., the amount and type of fluids used, the timing of initiation and the dosing of vasopressors, which antibiotics are given, and whether to administer corticosteroids. 

In this work, we propose a data-driven approach to discover optimal sepsis treatment strategies. We use deep reinforcement learning (RL) algorithms to identify how best to treat septic patients in the intensive care unit (ICU) to improve their chances of survival. 
While RL has been used successfully in complex decision making tasks \citep{atari,go}, its application to clinical models has thus far been limited by data availability \citep{nemati} and the inherent difficulty of defining clinical state and action spaces \citep{prasad2017reinforcement,komorowski}. 

Nevertheless, RL algorithms have many desired properties for the problem of deducing high-quality treatments. Their intrinsic design for sparse reward signals makes them well suited to overcome complexity from the stochasticity in patient responses to medical interventions, and delayed indications of efficacy of treatments. Importantly, RL algorithms also allow us to infer optimal strategies from suboptimal training examples. 

In this work, we demonstrate how to surmount the modeling challenges present in the medical environment and use RL to successfully deduce optimal treatment policies for septic patients.\footnote{Either patients who develop sepsis in their ICU stay, or those who are already septic at the start of their stay.} We focus on continuous state-space modeling, represent a patient's physiological state at a point in time as a continuous vector (using either raw physiological data or sparse latent state representations), and find optimal actions with Deep-Q Learning \citep{atari}. Motivating this approach is the fact that physiological data collected from ICU patients provide very rich representations of a patient's physical state, allowing for the discovery of interpretable and high-quality policies.  

In particular, we:
\begin{enumerate}[nosep]
\item Propose deep reinforcement learning models with continuous-state spaces, improving on earlier work with discretized models. 
\item Identify treatment policies that could improve patient outcomes, potentially reducing patient mortality in the hospital by 1.8 - 3.6\%, from a baseline mortality of 13.7\%.
\item Investigate the learned policies for clinical interpretability and potential use as a clinical decision support tool.
\end{enumerate} 

\section{Background and Related Work}
In this section we outline important reinforcement learning algorithms used in the paper and motivate our approach in comparison to prior work. 
\subsection{Reinforcement Learning}
Reinforcement learning (RL) models time-varying state spaces with a Markov Decision Process (MDP), in which at every timestep $t$ an agent observes the current state of the environment $s_t$, takes an action $a_t$ from the allowable set of actions $\cal A = \{$1$, \dots, M\}$, receives a reward $r_t$, and then transitions to a new state $s_{t+1}$. The agent selects actions at each timestep that maximize its expected discounted future reward, or \emph{return}, defined as $R_t = \sum_{t'=t}^{T} \gamma^{t'-t}r_{t'}$, where $\gamma$ captures the tradeoff between immediate and future rewards, and $T$ is the terminal timestep. The optimal action value function $Q^{*}(s,a)$ is the maximum discounted expected reward obtained after executing action $a$ in state $s$; that is, performing $a$ in state $s$ and proceeding optimally from this point onwards. More concretely, $Q^{*}(s,a) = \max_{\pi}\mathbb{E}[R_t | s_t = s, a_t = a, \pi]$, where $\pi$ --- also known as the \emph{policy} --- is a mapping from states to actions. The optimal value function is defined as $V^{*}(s)=\max_{\pi}\mathbb{E}[R_t|s_t = s,\pi]$, where we act according to $\pi$ throughout. 

In Q-learning, the optimal action value function is estimated using the Bellman equation, \newline $Q^{*}(s,a) = \mathop{\mathbb{E}}_{s'\sim T(s'|s,a)}[r + \gamma \max_{a'} Q^{*}(s',a')| s_t = s, a_t = a]$, where $T(s'|s,a)$ refers to the state transition distribution. Learning proceeds either with value iteration \citep{sutton} or by directly approximating $Q^{*}(s,a)$ using a function approximator (such as a neural network) and learning via stochastic gradient descent. Note that Q-learning is an \emph{off-policy} algorithm, as the optimal action-value function is learned with samples $<s,a,r,s'>$ that are generated to explore the state space. An alternative to Q-learning is the SARSA algorithm \citep{sarsa}; an on-policy method to learn $Q^{\pi}(s,a)$, which is the action-value function when taking action $a$ in state $s$ at time $t$, and then proceeding according to policy $\pi$ afterwards. 

In this work, the state $s_t$ is a patient's physiological state, either in raw form (as discussed in Section 3.2) or as a latent representation. The action space, $\mathcal{A}$, is of size 25 and is discretized over doses of vasopressors and IV fluids, two drugs commonly given to septic patients, detailed further in Section 3.3. The reward $r_t$ is $\pm R_\textit{max}$ at terminal timesteps and zero otherwise, with positive rewards being issued when a patient survives. At every timestep, the agent is trained to take an action $a_t$ with the highest Q-value, aiming to increase the chance of patient survival. 

\subsection{Reinforcement Learning in Health}
Much prior work in clinical machine learning has focused on supervised learning techniques for diagnosis \citep{dnn-skincancer} and risk stratification \citep{riskstrat-diabetes}. The incorporation of time in a supervised setting could be implicit within the feature space construction~\citep{hug2009icu,joshi2012prognostic}, or captured with multiple models for different timepoints ~\citep{fialho2013disease,ghassemi2014unfolding}. We prefer RL for sepsis treatment over supervised learning, because the ground truth of ``good'' treatment strategy is unclear in medical literature \citep{demise-egdt}. Importantly, RL algorithms also allow us to infer optimal strategies from training examples that do not represent optimal behavior. RL is well-suited to identifying ideal septic treatment strategies, because clinicians deal with a sparse, time-delayed reward signal in septic patients, and optimal treatment strategies may differ.

\cite{nemati} applied deep RL techniques to modeling ICU heparin dosing as a Partially Observed Markov Decision Process (POMDP), using both discriminative Hidden Markov Models and Q-networks to discover the optimal policy. Their investigation was made more challenging by the relatively small amount of available data. \cite{shortreed2011informing} learned optimal treatment policies for schizophrenic patients, and quantified the uncertainty around the expected outcome for patients who followed the policies. \cite{prasad2017reinforcement} use off-policy reinforcement learning algorithms to determine ICU strategies for mechanical ventilation administration and weaning, but focus on simpler learning algorithms and a heuristic action space. We experiment with using a sparse autoencoder to generate latent representations of the state of a patient, likely leading to an easier learning problem. We also propose neural network architectures that obtain more robust methods for optimal policy deduction.

Optimal sepsis treatment strategy was tackled most recently by \cite{komorowski}, using a discretized state and action-space to deduce optimal treatment policies for septic patients. Their work applied on-policy SARSA learning to fit an action-value function to the physician policy and value-iteration techniques to find an optimal policy \citep{sutton}. The optimal policy was then evaluated by comparing the Q-values that would have been obtained following chosen actions to the Q-values obtained by the physicians. We reproduce a similar model as our baseline, using related data pre-processing and clustering techniques. We additionally build on this approach by extending the results to the continuous domain, where policies are learned directly from the physiological state data, without discretization. We also propose a novel evaluation metric, different from ones used in \cite{komorowski}. We focus on in-hospital mortality instead of 90-day mortality (used in \cite{komorowski}) because of the other unobserved factors that could affect mortality in a 3-month timeframe.

\section{Data and Preprocessing}
\subsection{Cohort}
Data for these patients were obtained from the Multiparameter Intelligent Monitoring in Intensive Care (MIMIC-III v1.4) database \citep{mimic}, which is publicly available, and contains hospital admissions from approximately 38,600 adults (at least 15 years old). We extracted a cohort of patients fulfilling the Sepsis-3 criteria \citep{sepsis3}, and note that summary information about the populations is similar in sepsis survivors and mortalities (Table \ref{tab:cohort}).
\begin{table}[htbp]
 \centering 
 \begin{tabular}{|l|l||l|l||l|}\hline
  & \% Female & Mean Age & Hours in ICU & Total Population \\ \hline
   	Survivors & 43.6 & 63.4 & 57.6 & 15,583 \\ \hline
	Non-survivors & 47.0 & 69.9 & 58.8 & 2,315 \\ \hline
 \end{tabular}
  \caption{Comparison of cohort statistics for subjects that fulfilled the Sepsis-3 criteria.} 
  \label{tab:cohort} 
\end{table}

\subsection{Feature Preprocessing}
For each patient, we extracted relevant physiological parameters including demographics, lab values, vital signs, and intake/output events. Data were aggregated into windows of 4 hours, with the mean or sum being recorded (as appropriate) when several data points were present in one window.
Variables with excessive missingness were removed, and any remaining missing values were imputed with k-nearest neighbors, 
yielding a $47\times1$ feature vector for each patient at each timestep. Values exceeding clinical limits were capped, and capped data was normalized per-feature to zero mean and unit variance. See Appendix \ref{sec:appendix_features} for a full feature list. 

\subsection{Action Discretization}
We defined a $5\times5$ action space for the medical interventions covering the space of intravenous (IV) fluid (volume adjusted for fluid tonicity). and maximum vasopressor (VP) dosage in a given 4 hour window. The action space was restricted to these two interventions as both drugs are extremely important in the management of septic patients, but there is no agreement on when, and how much, of each drug to give \citep{demise-egdt}. We discretized the action space into per-drug quartiles based on all non-zero dosages of the two drugs, and converted each drug at every timestep  into an integer representing its quartile bin. We included a special case of no drug given as bin 0. This created an action representation of interventions as tuples of (total IV in, max VP in) at each time.

\section{Methods}
The challenge of applying RL to optimal medication dosing is that all available data are \emph{offline sampled}; that is, data are collected previously and models can only be fit to a retrospective dataset. In an RL context, this limits exploration of the state space in question, and makes learning the truly `optimal' policy difficult. This limitation motivates trying several different approaches, with varied modeling constraints, to determine the best medication strategy for patients. 

We focus on off-policy RL algorithms that learn an optimal policy through data that is generated by following an alternative policy. This makes sense for our problem because the available data are generated from a policy followed by physicians, but our goal is to learn a different, optimal policy rather than to evaluate the physician's policy. We propose deep models with continuous state spaces and discretized action spaces to retain more of the underlying state representation. 

\subsection {Discretized State-space and Discretized Action-space}
Following \cite{komorowski}, we create a baseline model with discretized state and action spaces, aiming to capture the underlying representation while simplifying the learning procedure. We use this approach to evaluate the performance of other techniques, and to understand the significance of learned Q values. We use the SARSA algorithm \citep{sarsa} to learn $Q^{\pi}(s,a)$, and the action-value function for the physician policy (more detail in Appendix \ref{sec:appendix_train}).

\subsection {Continuous State-spaces}
Continuous state-space models directly capture a patient's physiological state, and allow us to discover high-quality treatment policies.  To learn an optimal policy with continuous state vectors, we use neural networks to approximate the optimal action-value function, $Q^{*}(s,a)$.

\subsubsection {Model Architecture}
\label{sec:continuous_spaces}
Our model is based on a variant of Deep Q Networks \citep{atari}. Deep Q Networks seek to minimize a squared error loss between the output of the network, $Q(s,a;\theta)$, and the desired target, $Q_{\textit{target}} = r + \gamma \max_{a'}Q(s',a';\theta)$, observing tuples of the form $<s,a,r,s'>$. The network has outputs for all the different actions that can be taken --- for all $a \in \cal A = \{$1$, \dots, M\}$. Concretely, the parameters $\theta^{*}$ are found such that:
$$\theta^{*} = \argmin_{\theta}\mathbb{E}\left[\mathcal{L}(\theta)\right] = \argmin_{\theta}\mathbb{E}\left[\left( Q_{\textit{target}} - Q(s,a;\theta)\right)^{2}\right] $$

In practice, the expected loss is minimized via stochastic batch gradient descent. However, this method can be unstable due to non-stationarity of the target values, and using a separate network to determine the target Q values ($Q(s',a')$), which is periodically updated towards the main network (used to estimate $Q(s,a)$) helps to improve performance. 

Simple Q-Networks have several shortcomings, so we made several important modifications to make our model suitable for this situation. Firstly, Q-values are frequently overestimated in practice, leading to incorrect predictions and poor policies. We solve this problem with a Double-Deep Q Network \citep{ddqn}, where the target Q values are determined using actions found through a feed-forward pass on the main network, as opposed to being determined directly from the target network. 
In the context of finding optimal treatments, we want to separate the influence on Q-values of 1) a patient's \textit{underlying state} being good (e.g. near discharge), and 2) the correct action being taken at that timestep. To this end, we use a Dueling Q Network \citep{dueling}, where the action-value function for a given $(s,a)$ pair, $Q(s,a)$,  is split into separate  \emph{value} and \emph{advantage} streams, where the \emph{value} represents the quality of the current state, and the \emph{advantage} represents the quality of the chosen action. 
Training such a model can be slow as reward signals are sparse and only available on terminal timesteps. We use Prioritized Experience Replay \citep{per} to accelerate learning by sampling a transition from the training set with probability proportional to the previous error observed. 

Our final network architecture is a Dueling Double-Deep Q Network (Dueling DDQN), combining both of the above ideas. The network has two hidden layers of size 128, uses batch normalization \citep{batchnorm} after each, Leaky-ReLU activation functions, a split into equally sized advantage and value streams, and a projection onto the action space by combining these two streams. For more details, see Appendix \ref{sec:appendix_model}. 

After training the Dueling DDQN, we can then obtain the optimal policy for a given patient state as: $\pi^{\ast}(s) = \argmax_{a} Q(s,a)$.

\subsection{Autoencoder Latent State Representation}
Deep RL approaches for optimal medication are challenging to learn, because patient state is a high-dimensional continuous vector without clear structure. We examined both ordinary autoencoders \citep{autoencoder} and sparse autoencoders \citep{sparse-autoencoder} to produce latent state representations of the physiological state vectors and simplify the learning problem. Sparse autoencoders were trained with an additional term in the loss function to encourage sparsity. 
Our autoencoder models all had a single hidden layer, which was used as the latent state representation. These latent state representations were used as inputs to the Dueling DDQN (Section \ref{sec:continuous_spaces}).

\section{Evaluation}
The evaluation of off-policy models is challenging, because it is difficult to estimate whether the rollout of a learned policy (using the learned policy to determine actions at each state) would eventually lead to lower patient mortality. Furthermore, directly comparing Q values on off-policy data, as done in prior applications of RL to healthcare \citep{komorowski} can provide incorrect performance estimates \citep{off-policy-eval}.  Finding the average Q-value as in \cite{komorowski} is suboptimal because the $Q^{\pi}$ used for assessment represents the expected return when acting optimally at state $s_t$, but then proceeding according to $\pi_{\textit{physician}}$, the physician policy.  
In this work, we propose evaluating learned policies with several approaches.

\subsection{Discounted Returns vs. Mortality} 
\label{sec:ev_discounted}
To understand how expected discounted returns relate to mortality, we bin Q-values obtained via SARSA on the test set into discrete buckets, and for each, if it is part of a trajectory where a patient died, we assign it a label of 1. If the patient survived, we assign a label of 0. These labels represent the ground truth, as we know the actual outcome of patients when the physician's policy is followed. We compute the average mortality in each bin, enabling us to produce an empirically derived function of proportion of mortality versus expected return (Figure \ref{fig:m_vs_r}). We expect to see an inverse relationship between mortality and expected return, and this function enables us to associate returns with mortality for the purpose of evaluation. 
%

\subsection{Off-Policy Evaluation}
\label{sec:ev_off}
We use the method of Doubly Robust Off-policy Value Evaluation \citep{off-policy-eval} to obtain an unbiased estimate of the value of the learned optimal policy using off-policy sampled data (the trajectories in our training set). For each trajectory $H$ we compute an unbiased estimate of the value of the learned policy, $V_{\textit{DR}}^{H}$,
and average the results obtained across the observed trajectories. More details are provided in \cite{off-policy-eval}. We can also compute the mean discounted return of chosen actions under the physician policy. 
Using both these estimates, and the empirically learned proportion of mortality vs.\ expected return function, we can assess the potential improvement our policy could bring in terms of reduction in patient mortality. This method allows to accurately compare the returns obtained via different methodologies on off-policy data and estimate the mortality we would observe when following the learned policies. Directly comparing returns without the use of such an estimator is likely to give invalid results \citep{off-policy-eval}.

\subsection{Qualitative Examination of Treatment Policies}
\label{sec:ev_qual}
We examine the overall choice of treatments proposed by the optimal policy to derive more clinical understanding, and compare these choices to those made by physicians to understand how differences in the chosen actions contribute to patient mortality.

\section{Results} 
\subsection {Fully Discretized Models are Well-calibrated with Test Set Mortality}
\begin{figure}[ht]
 \centering 
 \includegraphics[width=2.5in]{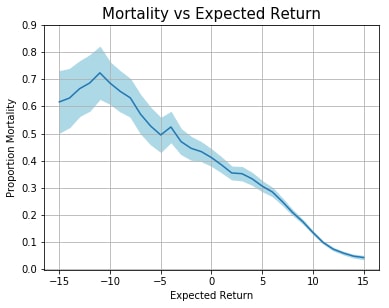} 
 \caption{The relationship between expected returns ---learned from observational data and actions taken by actual physicians --- and the risk of mortality in the test set of 3,580 patients (see Sec \ref{sec:ev_discounted}). The model appears to be well calibrated, with an inverse relationship between return and mortality. This function is not monotonically decreasing for low returns due to there being fewer training examples in this regime.}
 \label{fig:m_vs_r} 
\end{figure}

Figure \ref{fig:m_vs_r} shows the proportion of mortality versus the expected return for the physician policy on the held out test set. Note that $R_{\textit{max}} = 15$ is the reward issued at terminal timesteps. As shown, we observe high mortality with low returns, and low mortality with high returns. 
We also confirm that the empirically derived mortality for the physician's policy matches the actual proportion of mortality in the test set. For the empirically derived mortality, we average the expected return for the physician on the test set to obtain $13.9 \pm 0.5 \% $. This matches the actual proportion of mortality on the test set ($13.7\%$). 

\subsection {Continuous State-space Models}
We present the results for the two proposed networks: the Dueling Double-Deep Q Network (Dueling DDQN) and the Sparse Autoencoder Dueling DDQN. These are referred to as the \emph{normal Q-N} model and \emph{autoencode Q-N} model respectively for clarity. 

\subsubsection{Quantitative Value Estimate of Learned Policies}
\begin{figure}[htbp]
 \centering 
 \includegraphics[width=2.5in]{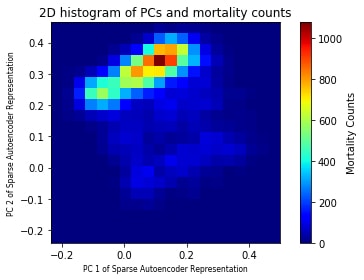} 
 \caption{Histogram of mortality counts against first two principal components of sparse autoencoder representation. Note the association between these values and the eventual outcome of the patient, potentially indicating why this model was able to learn a good quality policy.}
 \label{fig:heatmap} 
\end{figure}

Table \ref{tab:performance} demonstrates the relative performance of the three policies --- physician, \emph{normal Q-N}, and \emph{autoencode Q-N} --- on expected returns and estimated mortality. As described in Sec \ref{sec:ev_off}, we first obtain unbiased estimates of the value of our learned policies on the test data. The expected returns shown are $\bar{V}_{\textit{DR}}^{\textit{Physician}}$, $\bar{V}_{\textit{DR}}^{\textit{normal Q-N}}$, and $\bar{V}_{\textit{DR}}^{\textit{autoencode Q-N}}$. We estimate the mortality under each policy using Figure \ref{fig:m_vs_r}. As shown, the \emph{autoencode Q-N} policy has the lowest estimated mortality and could reduce patient mortality by up to 4\%. We examine a histogram of mortality counts against the first two principal components of the sparse representation (Figure \ref{fig:heatmap}) and observe a clear gradient of mortality counts, indicating how the autoencoder's hidden state may provide a rich representation of physiological state that leads to better policies.

\begin{table}[ht]
 \centering 
 \begin{tabular}{l|l||l}
    Policy & Expected Return & Estimated Mortality \\ \hline
  	Physician & 9.87 & $13.9 \pm 0.5 \%$ \\ \hline
   	Normal Q-N & 10.16 & $12.8 \pm 0.5 \%$ \\ \hline
	Autoencode Q-N & 10.73 & $11.2 \pm 0.4 \%$
 \end{tabular}
  \caption{Comparison of expected return and estimated mortality under the physician's policy, Normal Q-N, and Autoencode Q-N.} 
  \label{tab:performance} 
\end{table}


\subsubsection{Qualitative Examination of Learned Policies}
\begin{figure}[htbp]
 \centering 
 \includegraphics[width=6.5in]{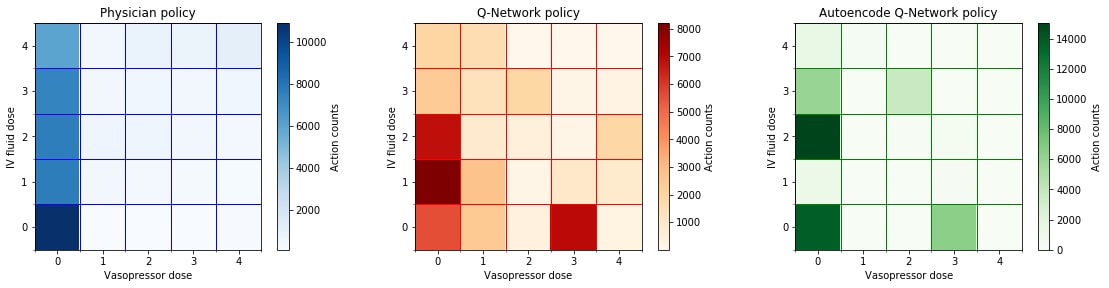} 
 \caption{Policies learned by the different models, as a 2D histogram, where we aggregate all actions selected by the physician and models on the test set over all timesteps. The axes labels index the discretized action space, where 0 represents no drug given, and 4 the maximum of that particular drug. Both models learn to prescribe vasopressors sparingly, a key feature of the physician's policy. }
 \label{fig:policies} 
\end{figure}

Figure \ref{fig:policies} demonstrates what the three policies --- physician, \emph{normal Q-N}, and \emph{autoencode Q-N} --- have learned as optimal policies. The action numbers index the different discrete actions selected at a given timestep, and the charts shown aggregate actions taken over all patient trajectories. Action 0 refers to no drugs given to the patient at that timestep, and increasing actions refer to higher drug dosages, where drug dosages are represented by quartiles. 

As shown, physicians do not often prescribe vasopressors to patients (note the high density of actions corresponding to vasopressor dose = 0) and this behavior is strongly in the policy learned by the \emph{autoencode Q-N} model. This result is sensible; even though vasopressors are commonly used in the ICU to elevate mean arterial pressure, many patients with sepsis are not hypotensive and therefore do not need vasopressors. In addition, there have been few controlled clinical trials that have documented improved outcomes from their use~\citep{mullner2004vasopressors}. The \emph{normal Q-N} also learns a policy where vasopressors are not given in with high frequency, but that policy is less evident. There are interesting parallels between the two learned policies (\emph{normal Q-N}, and \emph{autoencode Q-N}). For example, both favor action (0,2) (corresponding to no IV fluids given and an intermediate dosage of vasopressor given), and action (2,3) (corresponding to a medium dosage of IV fluids and vasopressors). 

\subsubsection{Quantifying Optimality of Learned Policies}
\begin{figure}[htbp!]
 \centering 
 \includegraphics[width=4in]{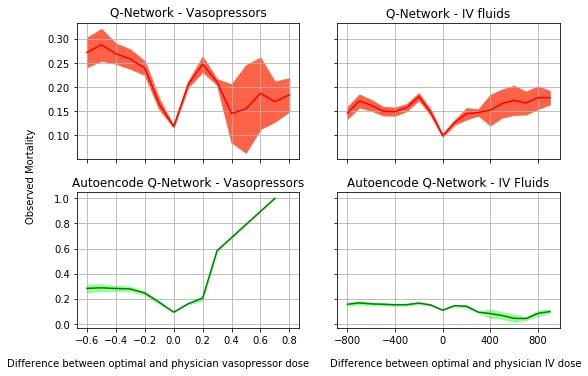} 
 \caption{Comparison of how observed mortality (y-axis) varies with the difference between the dosages recommended by the optimal policy and the dosages administered by clinicians (x-axis). For every timestep, this difference was calculated and associated with whether the patient survived or died in the hospital, allowing the computation of observed mortality. In general, we see low mortality for when the difference is zero, indicating that when the physician acts according to the optimal policy we observe more patient survival. }
 \label{fig:diff_mort} 
\end{figure}

Figure \ref{fig:diff_mort} shows the correlation between 1) the observed mortality, and 2) the difference between the optimal doses suggested by the policy, and the actual doses given by clinicians. The dosage differences at individual timesteps were binned, and mortality counts were aggregated. We observe consistently low mortalities when the optimal dosage and true dosage coincide, i.e. at a difference of 0, indicating the validity of the learned policy. The observed mortality proportion then increases as the difference between the optimal dosage and the true dosage increases. Results are less reliable when the optimal dose and physician dose differ by larger amounts. 

Both models appear to learn useful policies for vasopressors, with a large increase in observed mortality seen in the \emph{autoencode Q-N} because of relatively few cases in the test set where the optimal dose and given dose differed positively by a large amount. For IV-fluids, \emph{normal Q-N} learns a policy that shows a clear improvement over that of the physician's, indicated by the significant drop in observed mortality at the 0 mark. The \emph{autoencode Q-N} model learns a weaker policy over IV fluids, shown by the observed mortality decreasing as the difference between dosages increases.

\section{Conclusion}
In this work, we explored methods of applying deep reinforcement learning (RL) to the problem of deducing optimal medical treatments for patients with sepsis. There remain many interesting areas to be investigated. Firstly, the credit assignment in this model is quite sparse, with rewards/penalties only being issued at terminal states. There is scope for improvement here; one idea could be to use a clinically informed reward function based on patient blood counts to help learn better policies. Another approach could be to use inverse RL techniques \citep{Abbeel2010} to derive a suitable reward function based on the actions of experts (the physicians). As our dataset of patient trajectories is collected from recording the actions of many different physicians, this approach may allow us to infer a more appropriate reward function and in turn learn a better model.

Our contributions build on recent work by \cite{komorowski}, investigating a variety of techniques to find optimal treatment policies that improve patient outcome. We started by building a discretized state and action-space model, where the underlying states represent the physiological data averaged over four hour blocks and the action space is over two commonly administered drugs for septic patients --- IV fluids and vasopressors. Following this, we explored a fully continuous state-space/discretized action-space model, using Dueling Double-Deep Q Networks to learn an approximation for the optimal action-value function, $Q^{*}(s,a)$. 

We demonstrated that using continuous state space modeling found policies that could reduce patient mortality in the hospital by 1.8--3.6\%, which is an exciting direction for identifying better medication strategies for treating patients with sepsis. 
Our policies learned that vasopressors may not be favored as a first response to sepsis, which is sensible given that vasopressors may be harmful in some populations~\citep{d2015blood}. Our learned policy of intermediate fuild dosages also fits well with recent clinical work finding that large fluid dosages on first ICU day are associated with increased hospital costs and risk of death~\citep{marik2017fluid}. The learned policies are also clinically interpretable, and could be used to provide clinical decision support in the ICU. To our knowledge, this is the first extensive application of novel deep reinforcement learning techniques to medical informatics, building significantly on the findings of \cite{nemati}.

\acks{This research was funded in part by the Intel Science and Technology Center for Big Data and the National Library of Medicine Biomedical Informatics Research Training grant (NIH/NLM 2T15 LM007092-22). }

\newpage
\bibliography{writeup}
\newpage
\section{APPENDICES}

\subsection{Cohort definition}

Following the latest guidelines, sepsis was defined as a suspected infection (prescription of antibiotics and sampling of bodily fluids for microbiological culture) combined with evidence of organ dysfunction, defined by a Sequential Organ Failure Assessment (SOFA) score greater or equal to 2 \citep{sepsis3}. We assumed a baseline SOFA of zero for all patients. For cohort definition, we respected the temporal criteria for diagnosis of sepsis: when the microbiological sampling occurred first, the antibiotic must have been administered within 72 hours, and when the antibiotic was given first, the microbiological sample must have been collected within 24 hours \citep{sepsis3}. The earliest event defined the onset of sepsis. We excluded patients who received no intravenous fluid, and those with missing data for 8 or more out of the 47 variables. This method yield a cohort of 17,898 patients.

\subsection{Data extraction}

MIMIC-III was queried using pgAdmin 4. Raw data were extracted for all 47 features and processed in Matlab (version 2016b). Data was included from up to 24h preceding until 48h following the onset of sepsis, in order to capture the early phase of its management including initial resuscitation, which is the time period of interest. The features were converted into multidimensional time series with a time resolution of 4 hours. The outcome of interest was in-hospital mortality.


\subsection {Model Features}

\label{sec:appendix_features}
The physiological features used in our model are presented below. \newline\newline
\textbf{Demographics/Static}\newline
Shock Index, Elixhauser, SIRS, Gender, Re-admission, GCS - Glasgow Coma Scale, SOFA - Sequential Organ Failure Assessment, Age \newline \newline
\textbf{Lab Values}\newline
Albumin, Arterial pH, Calcium, Glucose, Haemoglobin, Magnesium, PTT - Partial Thromboplastin Time, Potassium, SGPT - Serum Glutamic-Pyruvic Transaminase, Arterial Blood Gas, BUN - Blood Urea Nitrogen, Chloride, Bicarbonate, INR - International Normalized Ratio, Sodium, Arterial Lactate, CO2, Creatinine, Ionised Calcium, PT - Prothrombin Time, Platelets Count, SGOT - Serum Glutamic-Oxaloacetic Transaminase, Total bilirubin, White Blood Cell Count \newline \newline
\textbf{Vital Signs} \newline
Diastolic Blood Pressure, Systolic Blood Pressure, Mean Blood Pressure, PaCO2, PaO2, FiO2, PaO/FiO2 ratio, Respiratory Rate, Temperature (Celsius), Weight (kg), Heart Rate, SpO2 \newline \newline
\textbf{Intake and Output Events}\newline
Fluid Output - 4 hourly period, Total Fluid Output, Mechanical Ventilation 

\subsection{Discretized State and Action Space Model}
\label{sec:appendix_train}
We present here how the discretized model was built.

\subsubsection {State Discretization}
As in the continuous case, the data are partitioned into a training set (80\%) and held-out test set (20\%) by selecting a proportionate number of patient trajectories for each set. These sets were checked to ensure they provide an accurate representation of the complete dataset, in terms of distribution of outcomes and some demographic features.
We apply k-means clustering to the training set, discretizing the states into 1250 clusters. As in \cite{komorowski}, we use a simple, sparse reward function, issuing a reward of +15 at a timestep if a patient survives, -15 if they die, and 0 otherwise. Test set data points are discretized according to whichever training set cluster centroid they fall closest to.

\subsubsection{SARSA for Physician Policy}
\label{sec:appendix_sarsa}
To learn the action-value function associated with the model, we used an offline, SARSA approach with the Bellman optimality equation, randomly sampling trajectories from our training set, and using tuples of the form $<s,a,r,s',a'>$ to update the action-value function:
\begin{center}
$Q(s, a) \leftarrow  Q(s, a) + \alpha \ast [r + \gamma Q(s', a')$ - $Q(s, a)]$
\end {center}
 Here, $(s,a)$ is the current (state, action) tuple considered, $(s',a')$ is a tuple representing the next state and action,  $\alpha$ is the learning rate and $\gamma$ the discount factor. As our state and action spaces are both finite in this model, we represent the Q-function using a table with rows for each $(s,a)$ tuple. This learned function was then used in model evaluation - after convergence, it represents $Q^{\pi}(s,a) = \mathop{\mathbb{E}}_{s'\sim T(s'|s,a)}[r + \gamma Q^{\pi}(s',a')| s_t = s, a_t = a, \pi]$, where $\pi$ is the physician policy.
 
\subsection {Continuous Model Architecture and Implementation Details}
\label{sec:appendix_model}
Our final network architecture had two hidden layers of size 128, using batch normalization \citep{batchnorm} after each, Leaky-ReLU activation functions, a split into equally sized advantage and value streams, and a projection onto the action space by combining these two streams together. 

The activation function is mathematically described by: $f(z) = \max(z,0.5z)$, where z is the input to a neuron.
This choice of activation function is motivated by the fact that Q-values can be positive or negative, and standard ReLU, tanh, and sigmoid activations appear to lead to saturation and `dead neurons' in the network. Appropriate feature scaling helped alleviate this problem, as did issuing rewards of $\pm15$ at terminal timesteps to help model stability.

We added a regularization term to the standard Q-network loss that penalized output Q-values which were outside of the allowed thresholds ($\pm15$), in order to encourage the network to learn a more appropriate Q function. Clipping the target network outputs to $\pm15$ was also found to be useful. The final loss function was: 
$$\mathcal{L}(\theta) = \mathbb{E}\left[\left(Q_{\textit{double-target}} - Q\left(s,a;\theta\right)\right)^{2}\right] + \lambda \cdot \max\left(\left| Q(s,a;\theta)-R_{\max} \right|,0\right)$$ 
with $R_{\max}$ being the absolute value of the reward/penalty issued at a terminal timestep, and 
$$ Q_{\textit{double-target}} = r + \gamma Q(s’,\argmax_{a'}Q(s’,a';\theta);\theta ')$$
where $\theta$ are the weights used to parameterize the main network, and $\theta '$ are the weights used to parameterize the target network.

We use a train/test split of 80/20 and ensure that a proportionate number of patient outcomes are present in both sets. Batch normalization is used during training. All models were implemented in TensorFlow v1.0, with Adam being used for optimization \citep{adam}.

During training, we sample transitions of the form $<s,a,r,s'>$ from our training set, perform feed-forward passes on the main and target networks to evaluate the output and loss, and update the weights in the main network via backpropagation.

\subsection{Autoencoder Implementation Details}
For the autoencoder, a desired sparsity $\rho$ is chosen, and the weights of the autoencoder are adjusted to minimize  ${\mathcal{L}_{\textit{sparse}}(\theta) = \mathcal{L}_{\textit{reconstruction}}(\theta) + \beta \sum_{j=1}^{n}\textit{KL}(\rho || \rho_{j})}$. Here, $n$ is the total number of hidden neurons in the network, $\rho_{j}$ is the actual output of neuron $j$, $\beta$ is a hyperparameter controlling the strength of the sparsity term, $\textit{KL}(\cdot || \cdot)$ is the KL divergence, and $\mathcal{L}_{\textit{reconstruction}}$ is the loss for a normal autoencoder. 

\end{document}